\documentclass[preprint,12pt]{elsarticle}

\usepackage{amssymb}
\usepackage{amsmath}

\usepackage{acro}
\usepackage{adjustbox}
\usepackage{booktabs}
\usepackage[hidelinks]{hyperref}
\usepackage{multirow}
\usepackage{subcaption}
\DeclareAcronym{af}{
  short = AF ,
  long  = Atrial Fibrillation
}

\DeclareAcronym{egms}{
  short = EGMs ,
  long  = Intracavitary electrograms
}

\DeclareAcronym{ml}{
  short = ML ,
  long  = machine learning
}

\DeclareAcronym{dl}{
  short = DL ,
  long  = deep learning
}

\DeclareAcronym{cae}{
  short = CAE ,
  long  = convolutional autoencoder
}

\DeclareAcronym{mse}{
  short = MSE ,
  long  = Mean Squared Error
}

\DeclareAcronym{tsne}{
  short = t-SNE ,
  long  = t-distributed Stochastic Neighbor Embedding
}

\DeclareAcronym{lr}{
  short = LR ,
  long  = Logistic Regression
}

\DeclareAcronym{knn}{
  short = KNN ,
  long  = k-Nearest Neighbors
}

\DeclareAcronym{auc}{
  short = AUC ,
  long  = area under the ROC curve
}

\begin{document}

\begin{frontmatter}



\title{Latent Representations of Intracardiac Electrograms for Atrial Fibrillation Driver Detection}


\author[Bioengineering]{Pablo Peiro-Corbacho} 
\author[Bioengineering]{Long Lin} 
\author[Hospital,CiberCV]{Pablo Ávila}
\author[Hospital,CiberCV]{Alejandro Carta-Bergaz}
\author[Hospital,CiberCV]{Ángel Arenal}
\author[SignalProcessing]{Carlos Sevilla-Salcedo\fnref{fn1}} 
\author[Bioengineering,Hospital,CiberCV]{Gonzalo R. Ríos-Muñoz\corref{cor1}\fnref{fn1}} 
\ead{grios@ing.uc3m.es}

\cortext[cor1]{Corresponding authors}
\fntext[fn1]{These authors contributed equally}

\affiliation[Bioengineering]{organization={Department of Bioengineering, Universidad Carlos III de Madrid},
            addressline={Av. de la Universidad, 30}, 
            city={Leganés},
            postcode={28911}, 
            state={Madrid},
            country={Spain}}

\affiliation[Hospital]{organization={Hospital General Universitario Gregorio Marañón, Instituto de Investigación Sanitaria Gregorio Marañón (IiSGM)},
            addressline={Dr. Esquerdo, 46}, 
            city={Madrid},
            postcode={28007}, 
            state={Madrid},
            country={Spain}}

\affiliation[CiberCV]{organization={CiberCV},
            addressline={Avenida Monforte de Lemos, 3-5}, 
            city={Madrid},
            postcode={28029}, 
            state={Madrid},
            country={Spain}}

\affiliation[SignalProcessing]{organization={Department of Signal Processing and
Communications, Universidad Carlos III de Madrid},
            addressline={Av. de la Universidad, 30}, 
            city={Leganés},
            postcode={28911}, 
            state={Madrid},
            country={Spain}}

\begin{abstract}

Atrial Fibrillation (AF) is the most prevalent sustained arrhythmia, yet current ablation therapies, including pulmonary vein isolation, are frequently ineffective in persistent AF due to the involvement of non-pulmonary vein drivers. This study proposes a deep learning framework using convolutional autoencoders for unsupervised feature extraction from unipolar and bipolar intracavitary electrograms (EGMs) recorded during AF in ablation studies. These latent representations of atrial electrical activity enable the characterization and automation of EGM analysis, facilitating the detection of AF drivers.

The database consisted of 11,404 acquisitions recorded from 291 patients, containing 228,080 unipolar EGMs and 171,060 bipolar EGMs. The autoencoders successfully learned latent representations with low reconstruction loss, preserving the morphological features. The extracted embeddings allowed downstream classifiers to detect rotational and focal activity with moderate performance (AUC 0.73–0.76) and achieved high discriminative performance in identifying atrial EGM entanglement (AUC 0.93).

The proposed method can operate in real-time and enables integration into clinical electroanatomical mapping systems to assist in identifying arrhythmogenic regions during ablation procedures. This work highlights the potential of unsupervised learning to uncover physiologically meaningful features from intracardiac signals.

\end{abstract}



\begin{keyword}

Atrial fibrillation \sep Intracavitary electrograms \sep Focal activity \sep Rotors \sep Entanglement \sep  Deep Learning \sep Convolutional Autoencoder



\end{keyword}

\end{frontmatter}



\section{Introduction}
\label{Sec:introduction}

\ac{af} is the most common sustained cardiac arrhythmia in adults, affecting an estimated 59 million people around the world in 2019 \cite{roth2020global}. It is defined as a supraventricular tachyarrhythmia characterized by disorganized electrical activity of the atrium and ineffective atrial contraction \cite{van20242024}. As life expectancy increases worldwide, the prevalence of \ac{af} is expected to rise accordingly \cite{linz2024atrial}. Although some patients may be asymptomatic, many experience symptoms such as palpitations, fatigue, and dyspnea. Besides, \ac{af} is associated with an elevated risk of heart failure, stroke, thromboembolic events, and cognitive decline \cite{hindricks20212020, buja2024cost}.

Mechanisms that initiate and perpetuate \ac{af} include complex spatio-temporal dynamics such as multiple wavelet propagation circuits, focal activity, and electrical vortices of self-sustained rotational activity (rotors), which have been particularly implicated in the maintenance of high-frequency activation \cite{wijesurendra2019mechanisms, lau2016novel, rios2018real}. Despite extensive research, the precise electrophysiological drivers of \ac{af}, particularly in persistent forms, remain a topic of debate, underscoring the need for deeper characterization of arrhythmogenic substrates \cite{van20242024}.

Catheter ablation has become a cornerstone therapy for rhythm control in \ac{af}. Pulmonary vein isolation (PVI) is the current standard approach, especially effective in paroxysmal \ac{af}, as pulmonary veins are recognized as frequent sources of ectopic activity \cite{tzeis20242024}. However, in persistent \ac{af}, ablation outcomes are suboptimal, reflecting the involvement of more complex and distributed arrhythmogenic sources beyond the pulmonary veins \cite{pak2019catheter}. This has led to intensified efforts to map and ablate non-pulmonary vein drivers.

\ac{egms}, acquired during electrophysiological studies using intracardiac catheters, provide localized high-resolution recordings of atrial electrical activity \cite{de2019electrogram}. These signals are pivotal for identifying regions of abnormal conduction and guiding ablation strategies. However, their interpretation remains highly dependent on the operators' expertise, and there is no universally accepted framework to automatically classify their underlying mechanisms.

To address this, we propose a feature extraction strategy based on \ac{dl}. Specifically, autoencoders are employed to learn compact, unsupervised representations (latent embeddings) of the morphology of the \ac{egms}. Unlike traditional methods based on hand-made features or linear transformations (e.g., statistical descriptors or PCA) \cite{singh2023ecg, greenacre2022principal}, autoencoders can capture non-linear dynamics and complex waveform patterns without prior assumptions \cite{chen2017deep, yildirim2018efficient, jang2021unsupervised}.

Recent work has begun to explore the use of \ac{ml} to classify atrial \ac{egms} and identify different \ac{af} mechanisms. Alhusseini et al. \cite{alhusseini2020machine} trained convolutional neural networks (CNNs) on phase maps from 35 persistent \ac{af} patients to detect rotational activity, achieving 96.1\% \ac{auc}. Rodrigo et al. \cite{rodrigo2022atrial} applied \ac{dl} to \ac{egms} from 86 patients, identifying \ac{af} signatures based on high activation rate, timing variability over 15\%, and morphological changes with beat-to-beat correlation below 0.48. Ríos-Muñoz et al. \cite{rios2022convolutional} trained a recurrent CNN on \ac{egms} from persistent \ac{af} patients to detect rotational activity, achieving 80\% accuracy and revealing that bipolar \ac{egms} better capture rotor-associated staircase patterns. Liao et al. \cite{liao2021deep} trained a residual CNN on unipolar \ac{egms} from 78 \ac{af} patients to detect focal sources, achieving 92.3\% \ac{auc} and highlighting periodic QS complexes critical to classification. Zolotarev et al. \cite{zolotarev2020optical} trained \ac{ml} models on Fourier features from MEM recordings in 11 human hearts, achieving 81\% F1-score for \ac{af} driver detection and improving localization by incorporating spatial neighborhood information.

However, few studies have focused on the extraction of low-dimensional latent spaces from \ac{egms} using unsupervised learning for mechanistic classification. Furthermore, existing models rarely address real-time applicability or generalize across the diversity of \ac{af} substrates.

In this work, we present a novel \ac{dl} pipeline for unsupervised representation learning of \ac{egms} recorded during atrial fibrillation. The core of our approach is a \ac{cae} that generates a latent space encoding relevant features of atrial activity. We evaluate the clinical interpretability of this latent space across three mechanistic classification tasks:
\begin{itemize}
    \item Detection of \textbf{focal activity} (localized high-frequency sources),
    \item Identification of \textbf{rotational activity} (reentrant drivers),
    \item Recognition of a new proposed \ac{af}-maintenance mechanism, \textbf{atrial entanglement}. This is defined as the acceleration of discrete \ac{egms} in neighboring sites concurrent with signal fragmentation in a given electrogram, being potentially indicative of functional reentry or mutual entrainment.
\end{itemize}

\begin{figure}[ht]
    \centering
    \includegraphics[width=\textwidth]{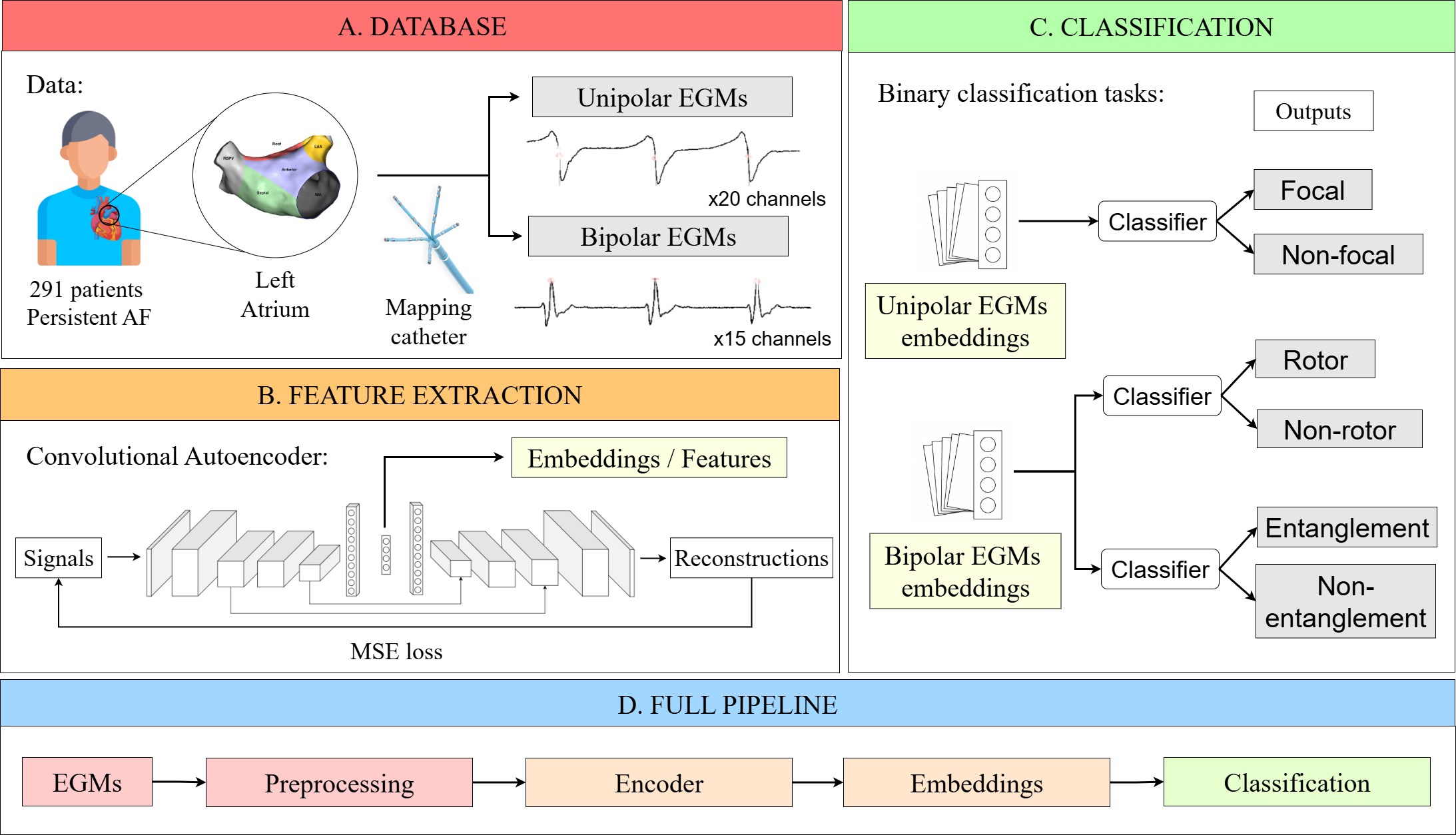}
    \caption{Proposed workflow. \textbf{A.} Bipolar and unipolar \ac{egms} are collected using a mapping catheter. \textbf{B.} \ac{cae} extracts latent embeddings from the \ac{egms}. \textbf{C.} Embeddings are then used as input for various classification models to perform binary classification of rotational, focal, and entanglement activity. \textbf{D. } Final pipeline of the study. AF, atrial fibrillation; EGM, electrogram; MSE, mean squared error.}
    \label{fig:workflow}
\end{figure}

This work aims to demonstrate that unsupervised latent spaces can encode mechanistically meaningful features that assist in the identification of \ac{af} driver regions. Ultimately, this could support the development of real-time AI-guided ablation strategies that extend beyond pulmonary vein isolation. A schematic representation of the proposed methodology is presented in Figure~\ref{fig:workflow}.

\section{Materials and Methods}
\label{Sec:materials}

\subsection{Database}
\label{Subsec:database}

The dataset comprised EGMs recorded from 291 patients with persistent AF who underwent catheter ablation at Hospital General Universitario Gregorio Marañón (Madrid, Spain). Table~\ref{tab:database description} summarizes the demographic and clinical characteristics of the cohort.

\begin{table}[ht!]
\centering
\begin{tabular}{lcc}
\toprule
\textbf{Characteristic} & \textbf{Value} & \textbf{Units} \\
\midrule
Age & 62.15 $\pm$ 9.03 & years \\
Sex (M/F) & 131 (75.29) / 43 (24.71)  & n (\%) \\
Body Mass Index (BMI) & 29.89 $\pm$ 4.52 & kg/m\textsuperscript{2} \\
Left atrial diameter & 7.71 $\pm$ 16.51 & mm \\
Left atrial area & 7.99 $\pm$ 12.05 & mm\textsuperscript{2} \\
Left ventricular ejection fraction & 52.89 $\pm$ 12.05 & \% \\
Previous cardioversion & 1.43 $\pm$ 1.09  & events \\
Hypertension & 100 (57.47) & n (\%) \\
Diabetes mellitus & 27 (15.52) & n (\%) \\
Heart failure & 66 (37.93) & n (\%) \\
Stroke & 14 (8.05) & n (\%) \\
OSA & 30 (17.24) & n (\%) \\
COPD & 8 (4.6) & n (\%) \\
NYHA & &  \\
\hspace{1em}I & 112 (38.49) & n (\%) \\
\hspace{1em}II & 58 (19.93)  & n (\%) \\
\hspace{1em}III & 3 (1.03) & n (\%) \\
\hspace{1em}IV & 1 (0.34) & n (\%) \\
\bottomrule
\end{tabular}
\caption{Database demographic and clinical statistics. These statistics were only available for 174 of the 291 patients (59.8\%). Values are mean ± std or n(\%), where percentages are calculated with respect to the number of patients for whom the statistics were available. OSA, Obstructive Sleep Apnea; COPD, Chronic Obstructive Pulmonary Disease; and NYHA, New York Heart Association classification.}
\label{tab:database description}
\end{table}

Simultaneous unipolar and bipolar EGMs were acquired using the CARTO3 system (Biosense Webster, CA, USA), an electroanatomical mapping platform that integrates catheter localization, real-time signal acquisition, and 3D anatomical reconstruction~\cite{narayan2024advanced}. Mapping was performed with a PentaRay catheter, consisting of five flexible splines arranged in a star configuration, each with four electrodes. This configuration allows for the simultaneous acquisition of 20 unipolar EGMs (one per electrode) and 15 bipolar EGMs (from adjacent electrode pairs), providing high spatial resolution of atrial electrical activity at each mapping site.

A total of 11,404 acquisitions were exported, yielding 228,080 unipolar and 171,060 bipolar EGMs, all sampled at 1000~Hz. Each signal lasted between 15 and 30 seconds. On average, $783.78 \pm 431.87$ unipolar and $587.84 \pm 323.90$ bipolar EGMs were recorded per patient.

\subsection{Driver Annotation Methods}
\label{Subsec:driver}



\begin{figure}[ht]
    \centering
    \includegraphics[width=\textwidth]{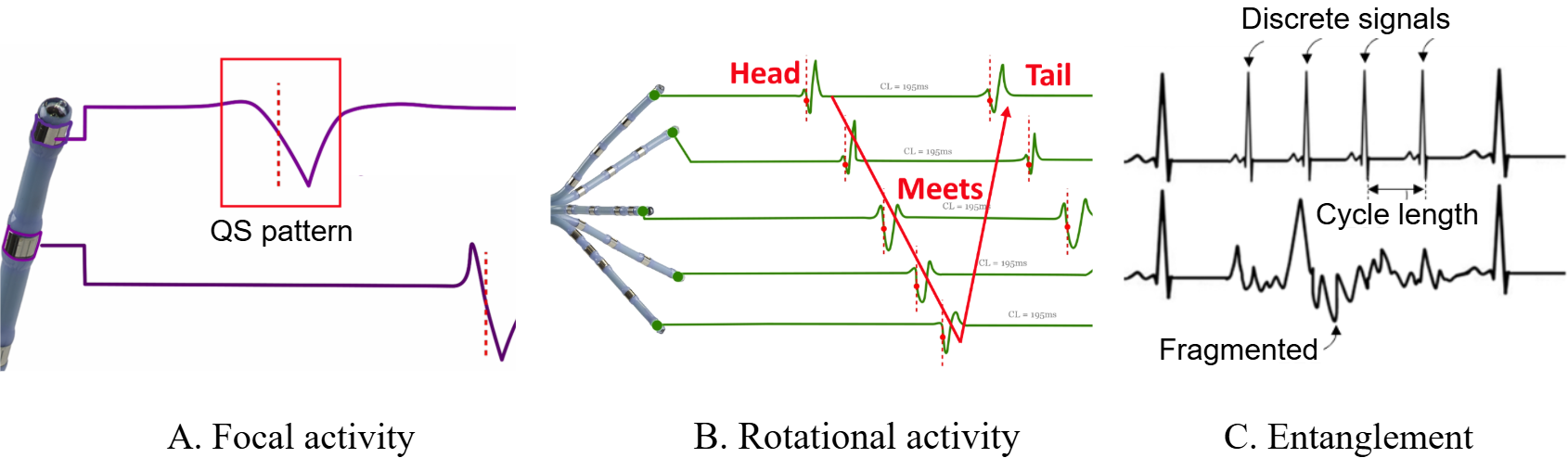}
    \caption{Graphical representation of how the different drivers are detected. \textbf{A.} QS patterns exhibiting focal activity. \textbf{B.}``Head meets tail" patterns present in rotational activity. \textbf{C.}~Shortening of cycle lengths duration of discrete activations localized in one electrode during a fragmented interval in another one, the EGM entanglement activity concept.}
    \label{fig:drivers}
\end{figure}

Figure~\ref{fig:drivers} illustrates the criteria used to identify each driver in the study. Driver annotations were derived using the CARTOFINDER module integrated within the CARTO3 electroanatomical mapping system. \ac{egms} exhibiting poor signal quality—assessed via amplitude thresholds, deflection sharpness, and signal-to-noise ratio—were excluded automatically.

\begin{itemize}

\item \textbf{Focal activity} was defined as repetitive, sharp QS deflections observed in unipolar \ac{egms}. Detection was performed within a 10~mm spatial radius and a 50~ms temporal window, identifying the earliest QS deflection. A region was annotated as focal if this pattern recurred over at least two consecutive beats (Figure~\ref{fig:drivers}A).

\item \textbf{Rotational activity} was identified from bipolar \ac{egms} using a circular electrode configuration—one electrode per spline, equidistant from the catheter center. Rotational patterns were defined by the presence of continuous head-meets-tail wavefronts covering more than 50\% of the dominant cycle length, with recurrence across at least two beats (Figure~\ref{fig:drivers}B).

\item \textbf{Entanglement activity} was annotated using the RAIS (Rotational Activation Identification System) in combination with an in-house software algorithm (patent pending). This method analyzed bipolar \ac{egms} to detect high-frequency complex activity (HFCA) and interspersed discrete intervals. An entanglement index was computed by evaluating cycle length shortening within these discrete intervals during HFCA episodes across neighboring \ac{egms}. Regions with an entanglement index greater than zero were classified as positive for entanglement (Figure~\ref{fig:drivers}C).

\end{itemize}

All annotated driver events were encoded as binary masks delineating their corresponding time intervals for subsequent use in model training and evaluation. 

\subsection{Signal preprocessing}
\label{Subsec:preprocessing}

\ac{egms} were originally recorded at 1000~Hz, yielding high-resolution signals of variable durations. To reduce dimensionality and computational load while preserving physiological information, all signals were resampled to 250~Hz~\cite{rios2022convolutional}. Unipolar (20 channels) and bipolar (15 channels) \ac{egms} were segmented into non-overlapping 1-second windows, resulting in fixed-size segments of 250 samples per channel. This ensured uniform input dimensions across the dataset. A total of 260,465 multi-channel segments were obtained from the 291 patients. Each segment was labeled as containing focal activity, rotational activity, or entanglement if it overlapped with any annotated driver event.

To prevent data leakage and ensure generalizability, the dataset was split by patient: 230 patients (207,864 segments, 79.8\%) for training, 27 patients (26,188 segments, 10.05\%) for validation, and 34 patients (26,413 segments, 10.14\%) for testing. This patient-wise split ensures that no subject contributes to more than one subset.

We applied signal normalization to reduce inter-patient variability and improve model convergence. Amplitude clipping was performed using the 2nd and 98th percentiles of the training data to suppress outliers. These thresholds were applied consistently to the validation and test sets. Following clipping, signals were min-max scaled to the range $[-1, 1]$ using:
\begin{equation}
    x' = 2 \cdot \frac{x - x_{\text{min}}}{x_{\text{max}} - x_{\text{min}}} - 1 ,
\end{equation}
where $x$ is the original value, $x'$ is the normalized value, and $x_{\text{min}}$, $x_{\text{max}}$ are the 2nd and 98th percentiles from the training set. Normalization parameters were computed exclusively from the training set to avoid information leakage. This step ensured stable training dynamics and consistent scaling across the dataset.

\subsection{Convolutional autoencoder} 

After preprocessing, each data instance was represented as a $15 \times 250$ matrix for bipolar \ac{egms} and $20 \times 250$ for unipolar \ac{egms}, where rows correspond to different channels and columns to time samples. These 2D representations were used as inputs to a \ac{cae} to learn compact, informative latent features.

The \ac{cae} consists of two main components: an encoder and a decoder. The encoder compresses the input signal $x$ into a low-dimensional latent representation $z = f_\theta(x)$ through a sequence of convolutional layers. The decoder reconstructs the original input from the latent space as $\hat{x} = g_\theta(z)$ using transposed convolutions and upsampling layers. The network is trained in a self-supervised manner, without external labels, to minimize reconstruction error~\cite{chen2017deep}.

This choice is motivated by several factors. The 2D structure preserves both temporal patterns and inter-signal relationships, allowing the model to learn from spatial dependencies. Convolutional layers are well-suited for capturing local features, enabling the \ac{cae} to detect recurring patterns in both time and across channels. Furthermore, weight sharing in convolutional layers leads to a model with fewer parameters, which helps reduce the risk of overfitting \cite{o2015introduction}. 

The implemented architecture takes advantage of key \ac{dl} techniques to ensure robust training. These techniques are the use of batch normalization to stabilize learning, dropout to improve generalization and reduce overfitting, and non-linear activations to capture complex patterns. In addition, a key technique that provided us with better results was the use of an encoder that applies max-pooling operations, which not only reduce spatial dimensions but also store pooling indices to guide the decoder in the reconstruction task. By reintroducing this spatial information during upsampling, the decoder can better preserve the overall structure, resulting in more coherent reconstructions. This design helps improve training efficiency and obtain better generalization performance.

The architecture is symmetric. The encoder consists of a series of convolutional blocks, each being a convolutional layer followed by batch normalization, ReLU activation, max pooling, and dropout, progressively reducing spatial dimensions. After the final encoder block, the feature maps are flattened and passed through a fully connected layer to produce a compact latent representation. The decoder mirrors this structure, starting with a fully connected layer whose output is reshaped to initiate the upsampling blocks. Each block performs max-unpooling using the stored indices, followed by transposed convolution, batch normalization, ReLU activation, and dropout, to gradually reconstruct the spatial resolution. The final layer applies a transposed convolution followed by a $tanh$ activation to match the input value range. Panel B in Figure~\ref{fig:workflow} presents a general sketch of the architecture.

In the context of autoencoders, the reconstruction loss measures how accurately the model can reproduce its input after encoding and decoding. A common choice for this purpose is the \ac{mse}, which calculates the average of the squared differences between each element of the original input and its corresponding reconstructed value. Formally, the \ac{mse} is defined as: 
\begin{equation}
    \mathcal{L}_{\text{MSE}} = \frac{1}{N} \sum_{i=1}^{N} (x_i - \hat{x}_i)^2,
\end{equation}
where $x_i$ and $\hat x_i$ represent the original and reconstructed values, respectively, and N is the number of samples. This loss function encourages the autoencoder to produce outputs that are as close as possible to the inputs. \ac{mse} directly reflects the quality of the reconstruction, making it a natural and interpretable objective for training the network.

For unipolar \ac{egms}, the standard \ac{mse} loss was sufficient to train the autoencoder effectively, resulting in accurate reconstructions. However, when applied to bipolar \ac{egms}, which are sparse signals, some of the models trained with \ac{mse} alone tended to simply predict zeros to minimize loss. To address this issue, we introduced a regularization term designed to penalize this behavior. Specifically, we developed a custom loss function that combines the standard \ac{mse} with an activity-aware regularization term to improve reconstruction performance by focusing more on active areas. This loss function is defined as:

\begin{equation}
    \mathcal{L}_{\text{regMSE}} = \mathcal{L}_{\text{MSE}} + \frac{\lambda_{\text{reg}}}{N} \cdot \frac{\sum_{i=1}^{N} \text{ReLU}(|y_i| - \tau) \cdot |y_i - \hat{y}_i|}{\sum_{i=1}^{N} \text{ReLU}(|y_i| - \tau) + \eta},
\end{equation}

where $y_i$ and $\hat y_i$ denote the true and predicted signal values, respectively, $\lambda_{reg}$ controls the strength of the regularization, $\tau$ defines the activity threshold, and $\eta$ is a small constant to prevent division by zero. This regularization term selectively penalizes reconstruction errors in activity regions, encouraging the model to focus on accurately reconstructing informative parts of the signal rather than minimizing errors in flat, inactive regions. Although this loss was effective on several models, the final proposed architecture did not benefit from it.

\subsection{Latent space visualization}

To assess the structure and interpretability of the learned latent representations, we applied \ac{tsne} to project the encoded features into two dimensions. t-SNE is a non-linear dimensionality reduction method that preserves local relationships by converting pairwise similarities into conditional probabilities and minimizing the Kullback–Leibler divergence between high- and low-dimensional distributions~\cite{maaten2008visualizing}.

This visualization enabled qualitative evaluation of the clustering behavior of the latent space. By plotting the \ac{tsne} outputs and color-coding by driver type, we examined whether the compressed representations retained class-specific information. The presence of separable clusters serves as evidence that the autoencoder captures meaningful, discriminative features relevant to downstream classification tasks.

\subsection{Classification head}

Latent embeddings generated by the autoencoder were used as input features for a set of machine learning classifiers to evaluate their ability to discriminate between different types of atrial activity. We considered several models with complementary properties: \ac{lr}, \ac{knn}, and three gradient boosting algorithms, i.e., XGBoost, LightGBM, and CatBoost.


Classification tasks were defined as three binary problems: detection of rotational, focal, and entanglement activity. As outlined in Section~\ref{Subsec:driver}, unipolar embeddings were used for focal activity detection, while bipolar embeddings were used for rotational and entanglement activity. Each task followed a one-vs-rest strategy: samples containing the target driver were labeled as positive (1), and all others, including different driver types and background, were labeled as negative (0).

Model performance was primarily evaluated using the \ac{auc}, which quantifies the trade-off between sensitivity and specificity across decision thresholds. Additional metrics, including accuracy, precision, and recall, were computed to provide a comprehensive assessment of classification performance.

\subsection{Hardware and software}

The models were implemented using Python 3.12.7. Training and evaluation were performed on a system equipped with an Intel® Core™ i7-9750H CPU @ 2.60 GHz (6 cores, 12 threads), 16 GB of RAM, and an NVIDIA GeForce RTX 2060 GPU with 6 GB of dedicated memory.

\section{Results}

\subsection{Experimental setup}

The autoencoder was implemented in PyTorch to leverage GPU acceleration during training. Optimization was performed using the Adam optimizer with a batch size of 128. Early stopping was applied based on the validation loss, halting training if the \ac{mse} did not improve by at least 0.00025 over five consecutive epochs.

\begin{figure}[ht]
    \centering
    \includegraphics[width=1\textwidth]{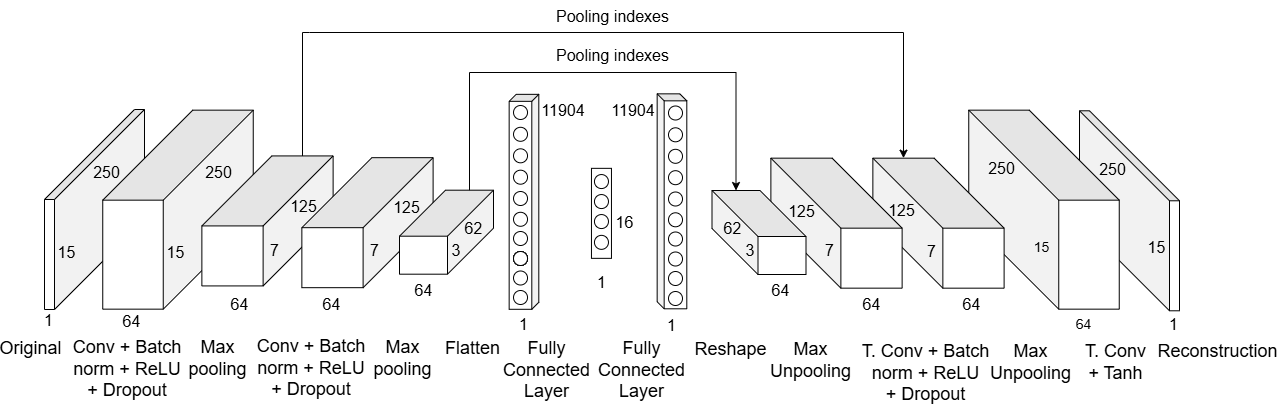}
    \caption{Architecture of the selected autoencoder used for extracting embeddings from bipolar \ac{egms}. The model consists of an encoder with convolutional, batch normalization, ReLU activation, and dropout layers, followed by max pooling and fully connected layers. The decoder mirrors this structure, incorporating unpooling, transposed convolutional layers, and a final reconstruction layer with Tanh activation. The same architecture is used on unipolar \ac{egms}, but changing data dimensions, as unipolar \ac{egms} have 20 channels instead of 15.}
    \label{fig:autoencoder_architecture}
\end{figure}

A grid search was conducted to optimize model performance, exploring the following hyperparameters: number of convolutional layers (1–3), number of filters per layer (16, 32, 64, 128), kernel size (3, 5), learning rate ($10^{-2}$, $10^{-3}$, $10^{-4}$), dropout rate (0.15, 0.25, 0.35), and latent space dimensionality (8, 16, 32, 64). The final configuration, selected based on validation loss and model simplicity, used two convolutional layers with 64 filters, kernel size 3, learning rate $10^{-3}$, dropout rate 0.25, and a latent dimension of 16. This architecture was adopted for both unipolar and bipolar  \ac{egms} and is illustrated in Figure~\ref{fig:autoencoder_architecture}. More complex architectures offered no substantial gain in performance and were discarded in favor of computational efficiency and reduced dimensionality.

For classification, latent embeddings were extracted from the same patient-based training, validation, and test splits used in the autoencoder training. Embeddings from the training and validation sets were combined and used in a 5-fold cross-validation process for model selection. The test set embeddings, derived from the autoencoder’s test patients, were reserved for final evaluation.

To address class imbalance, random undersampling was applied to the training data. The class distributions for each task are shown in Table~\ref{tab:activity}. All embeddings were standardized prior to training to ensure equal feature contribution.

\begin{table}[ht]
\centering
\begin{tabular}{lcccc}
\toprule
\multirow{2}{*}{\textbf{Activity Type}} & \multicolumn{2}{c}{\textbf{Training}} & \multicolumn{2}{c}{\textbf{Test}} \\
\cmidrule(lr){2-3} \cmidrule(lr){4-5}
 & \textbf{No activity} & \textbf{Activity} & \textbf{No activity} & \textbf{Activity} \\
\midrule
Rotational    & 3,471   & 3,471   & 26,124 & 289 \\
Focal         & 44,186  & 44,186  & 21,653     & 4,760 \\
Entanglement  & 47,015 & 47,015  & 5,393     & 21,020 \\
\bottomrule
\end{tabular}
\caption{Number of samples on the training and test sets for each classification task.}
\label{tab:activity}
\end{table}

Hyperparameters for the classification models were also tuned via grid search. For logistic regression, the regularization strength $C$ was varied across \{0.1, 1, 10\}. For \ac{knn}, the number of neighbors was selected from \{3, 5, 7, 11, 13, 15\}. For gradient boosting methods: XGBoost with number of estimators \{100, 200\}, max depth \{3, 6, 9\}; LightGBM with number of estimators \{100, 200\}, learning rate \{0.01, 0.1, 0.2\}; and CatBoost with number of estimators \{100, 200\}, max depth \{4, 6, 8\}. The best configuration for each classifier was selected based on the average \ac{auc} across validation folds.

\subsection{Model analysis}

We evaluated the performance of the autoencoders trained separately on unipolar and bipolar \ac{egms} using both quantitative metrics and qualitative inspection. The primary evaluation metric was reconstruction loss, complemented by visual comparison of original and reconstructed signals.

\begin{figure}[ht]
    \centering
    \begin{subfigure}{0.485\linewidth}
        \includegraphics[width=\linewidth]{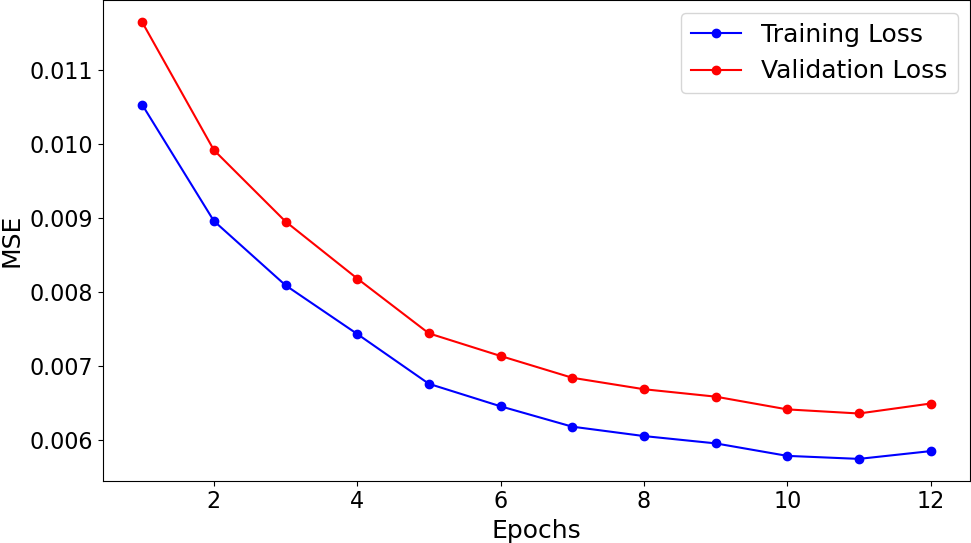}
        \caption{Bipolar \ac{egms}}
        \label{fig:loss_bipolar}
    \end{subfigure}
    ~
    \begin{subfigure}{0.485\linewidth}
        \includegraphics[width=\linewidth]{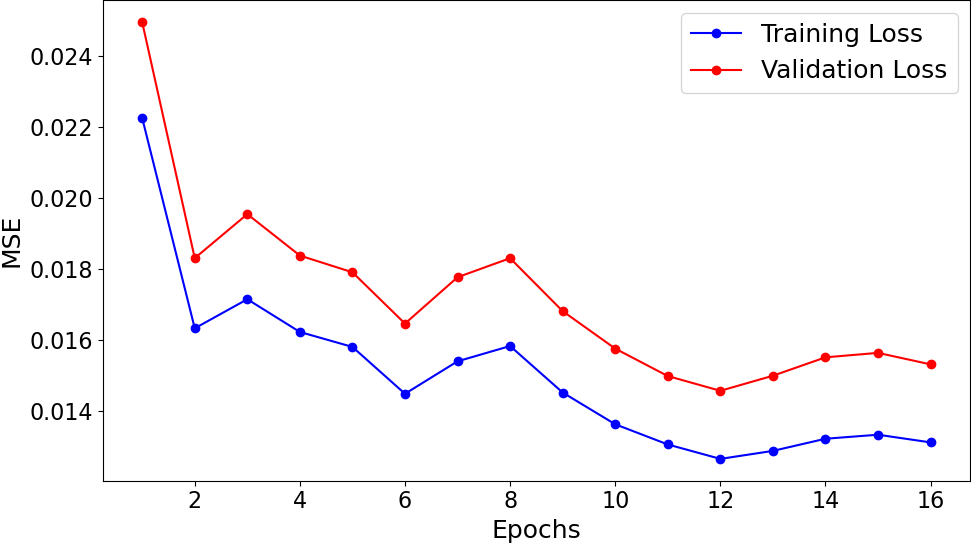}
        \caption{Unipolar \ac{egms}}
        \label{fig:loss_unipolar}
    \end{subfigure}
    \caption{Training and validation reconstruction losses obtained during the training of the convolutional autoencoders. EGM, electrogram; MSE, mean squared error. 
    }
    \label{fig:loss_plots}
\end{figure}

Figure~\ref{fig:loss_bipolar} displays the training and validation loss curves for the bipolar \ac{egms} autoencoder. Over 12 epochs, both losses decreased monotonically, indicating stable and effective learning. The narrow gap between curves suggests strong generalization with minimal overfitting. A slight increase in validation loss in the final epoch signals the onset of overfitting, validating the early stopping criterion, which selected the model from epoch 11. At this point, the training and validation losses reached 0.0057 and 0.0064, respectively.

Figure~\ref{fig:loss_unipolar} shows the loss curves for the unipolar \ac{egms} autoencoder. While minor fluctuations are observed across the 16 training epochs, the overall trend reflects consistent improvement. The moderate gap between training and validation losses indicates reasonable generalization. A slight divergence after epoch 12 marks the onset of overfitting, and the model from epoch 12 was selected, yielding training and validation losses of 0.0126 and 0.0146, respectively.

On the test set, the bipolar model achieved a reconstruction \ac{mse} of 0.0062, while the unipolar model achieved 0.0118. These low errors confirm that both models effectively preserve signal morphology in their compressed latent representations. The dimensionality was reduced from $15 \times 250$ (bipolar) and $20 \times 250$ (unipolar) to 16 latent features, demonstrating efficient compression without significant information loss.

\begin{figure}[ht!]
    \centering
    \begin{subfigure}{0.9\linewidth}
        \includegraphics[width=\linewidth]{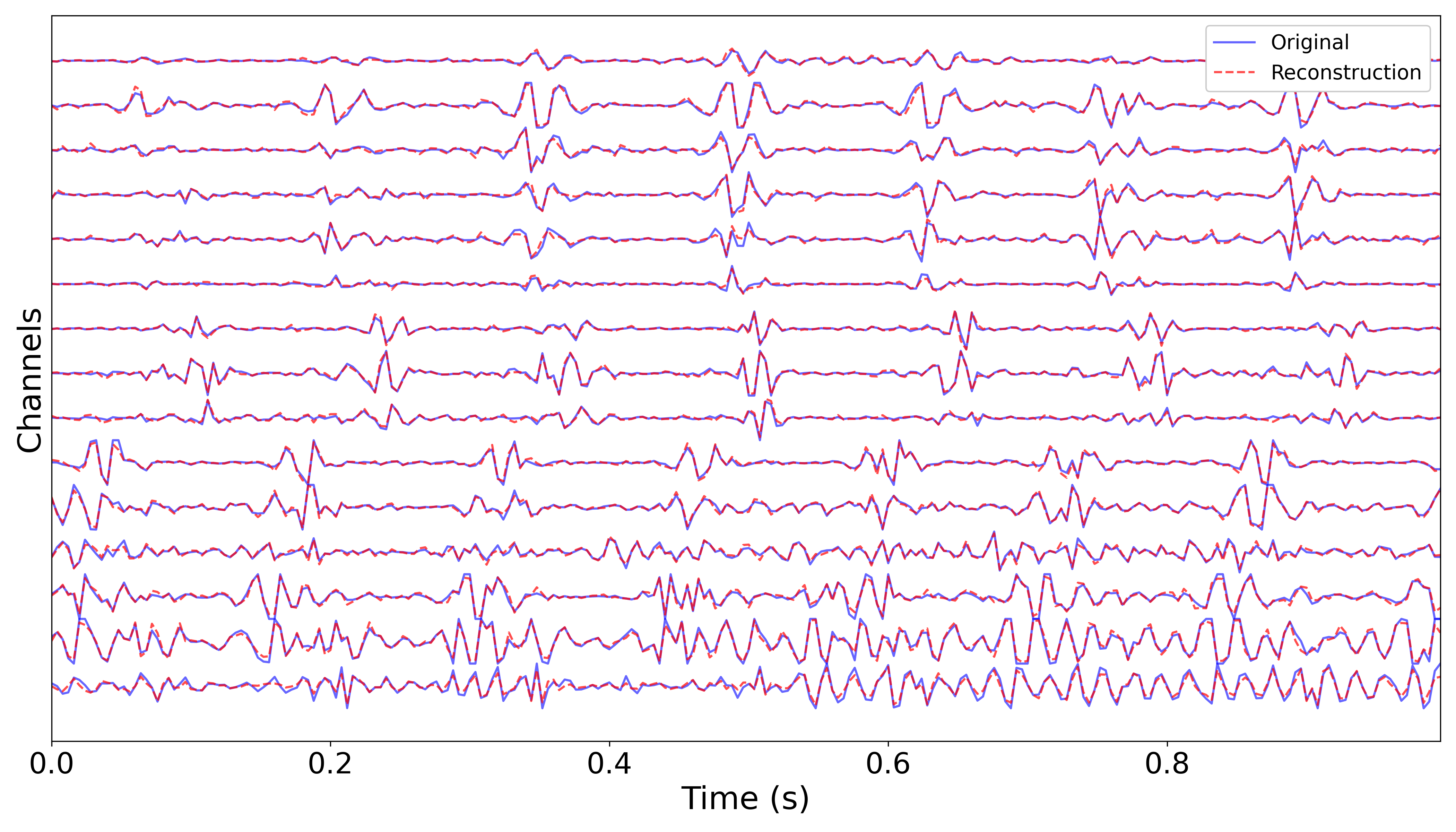}
        \caption{Bipolar \ac{egms}}
        \label{fig:reconstruction_bipolar}
    \end{subfigure}
    ~
    \begin{subfigure}{0.9\linewidth}
        \includegraphics[width=\linewidth]{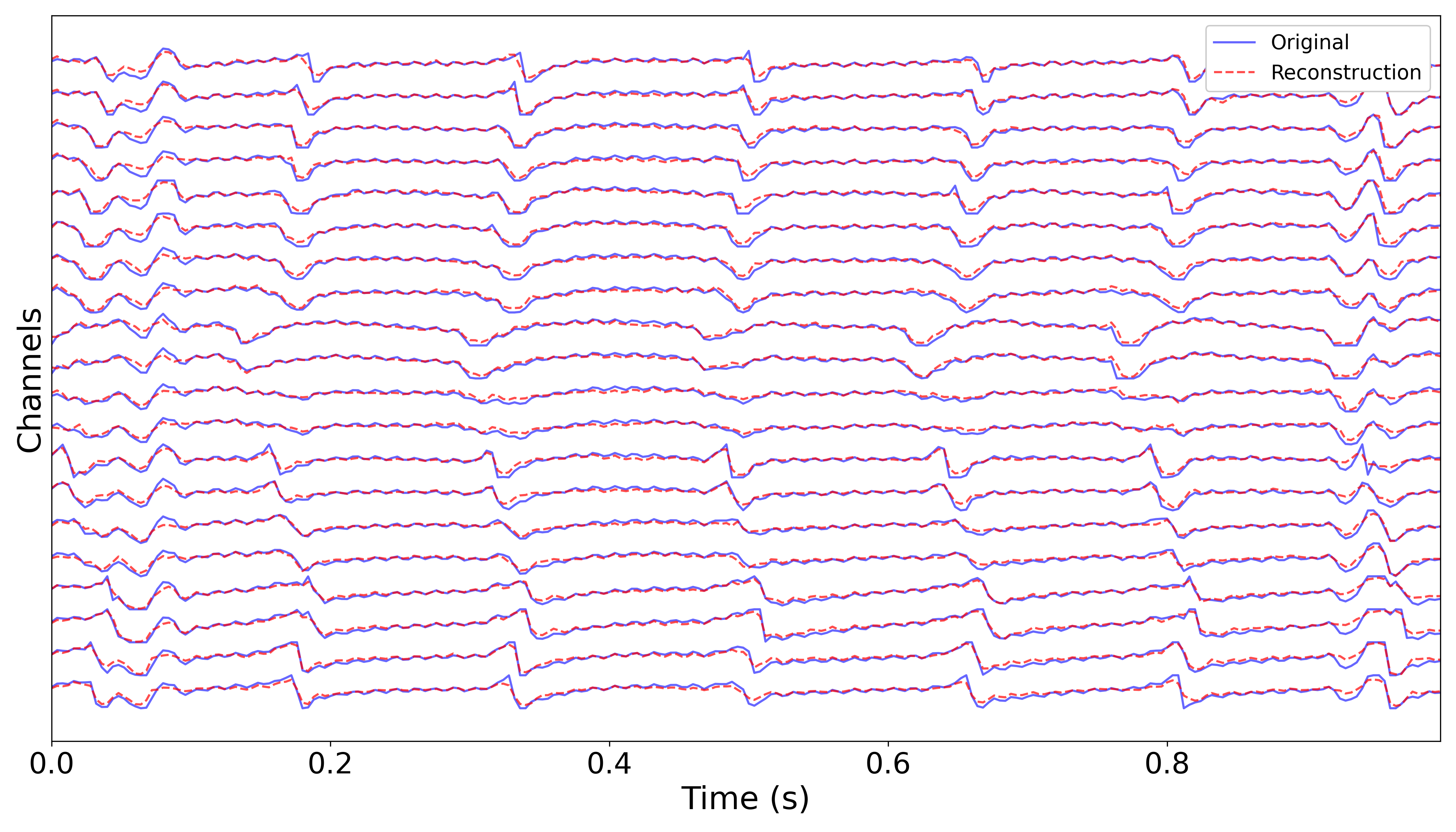}
        \caption{Unipolar \ac{egms}}
        \label{fig:reconstruction_unipolar}
    \end{subfigure}
    \caption{Reconstruction performance of the autoencoder on \ac{egms} from the test set. The blue solid lines represent the original signals, while the red dashed lines denote the reconstructed outputs from the model.
    }
    \label{fig:reconstructions}
\end{figure}

Figure~\ref{fig:reconstructions} illustrates representative reconstruction examples for bipolar (Figure~\ref{fig:reconstruction_bipolar}) and unipolar (Figure~\ref{fig:reconstruction_unipolar}) \ac{egms}. In both cases, the reconstructions closely match the original signals, with accurate preservation of waveform shape, peak timing, and amplitude. These results confirm the autoencoders' ability to learn meaningful low-dimensional representations suitable for downstream analysis.

\subsection{Classification performance analysis}

\begin{figure}[ht]
    \centering
    \begin{subfigure}{0.32\textwidth}
        \includegraphics[width=\linewidth]{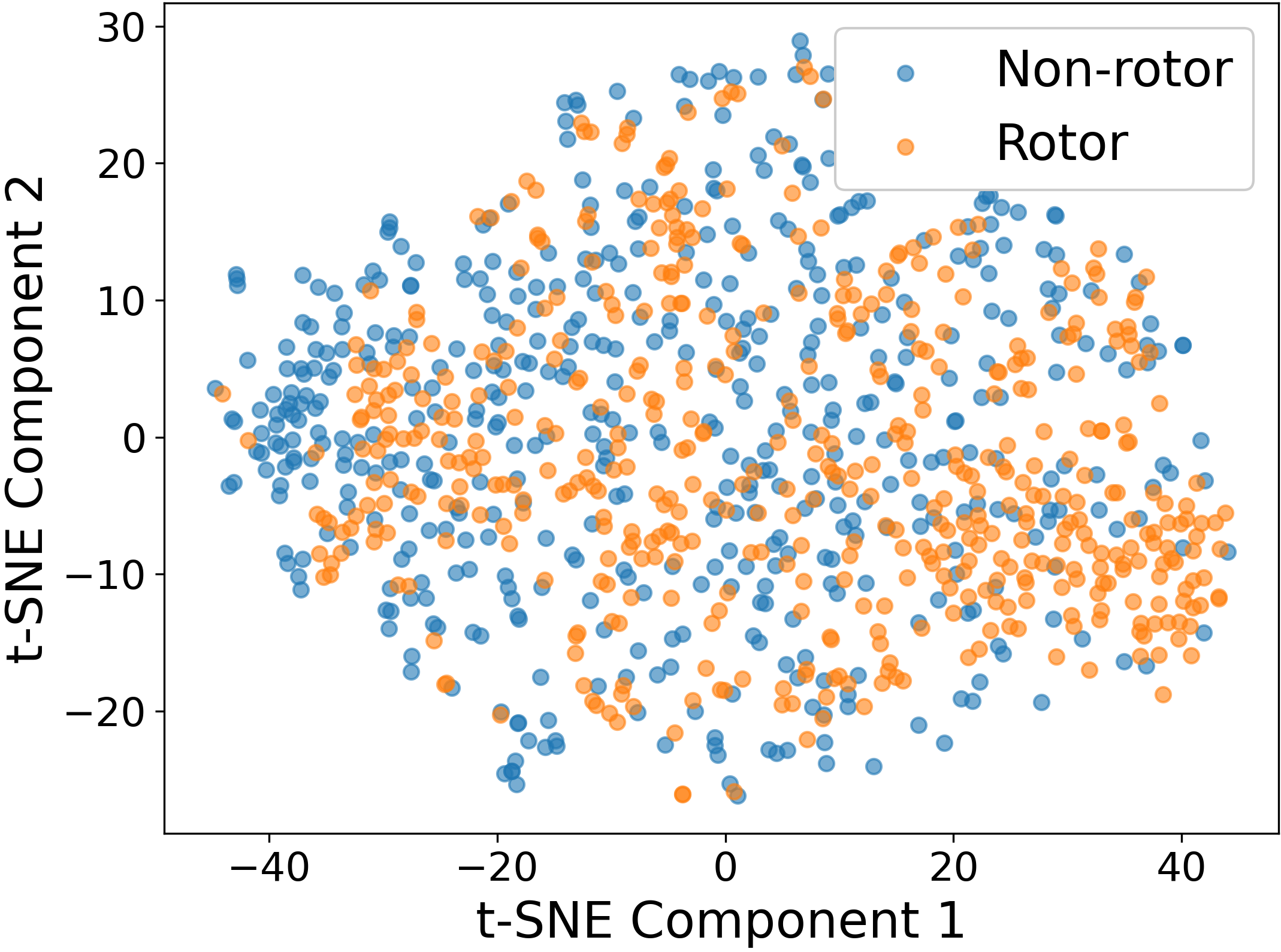}
        \caption{t-SNE - Rotor activity}
        \label{fig:tsne_rotor}
    \end{subfigure}
    \hfill
    \begin{subfigure}{0.32\textwidth}
        \includegraphics[width=\linewidth]{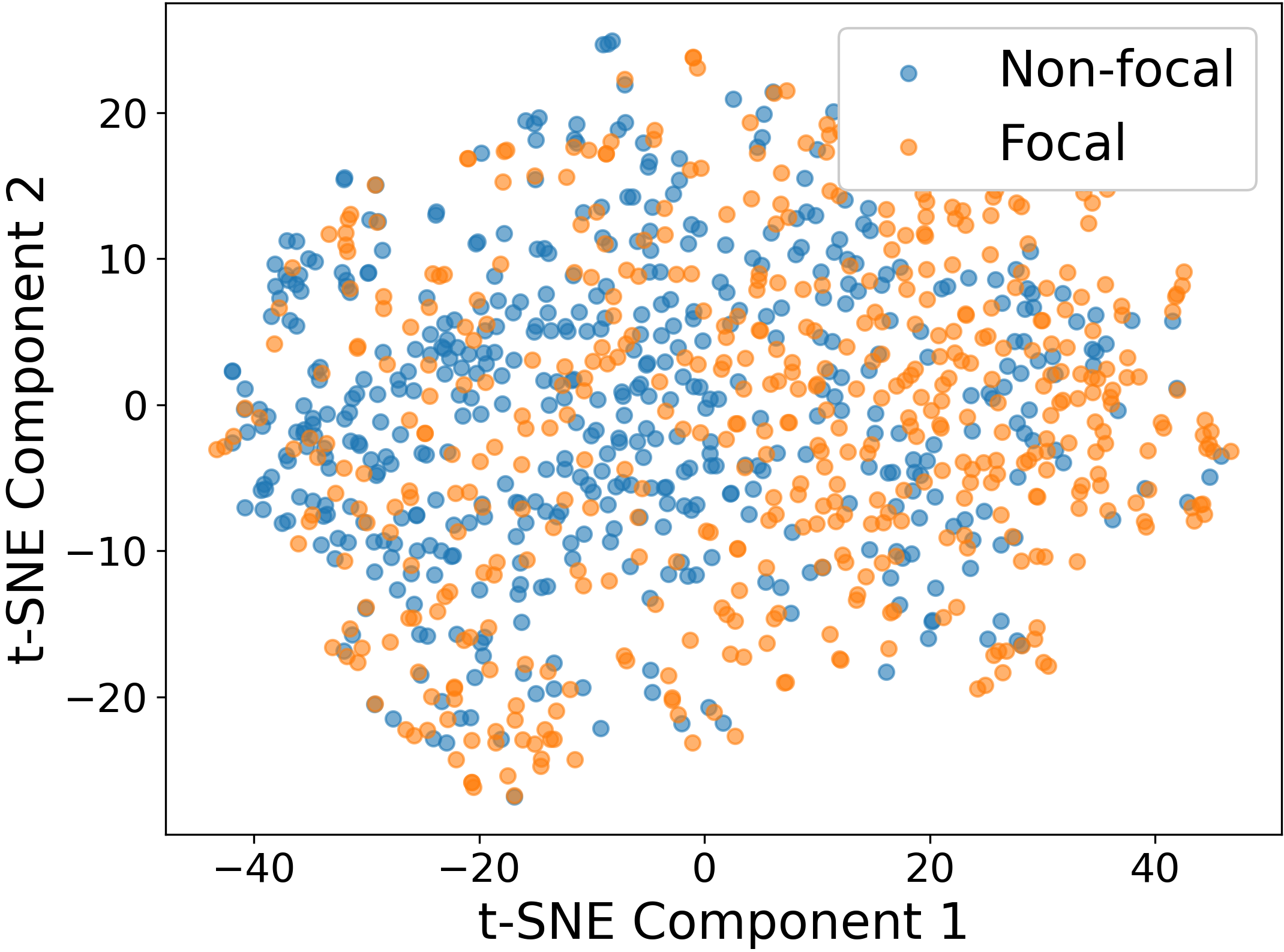}
        \caption{t-SNE - Focal activity}
        \label{fig:tsne_focal}
    \end{subfigure}
    \hfill
    \begin{subfigure}{0.32\textwidth}
        \includegraphics[width=\linewidth]{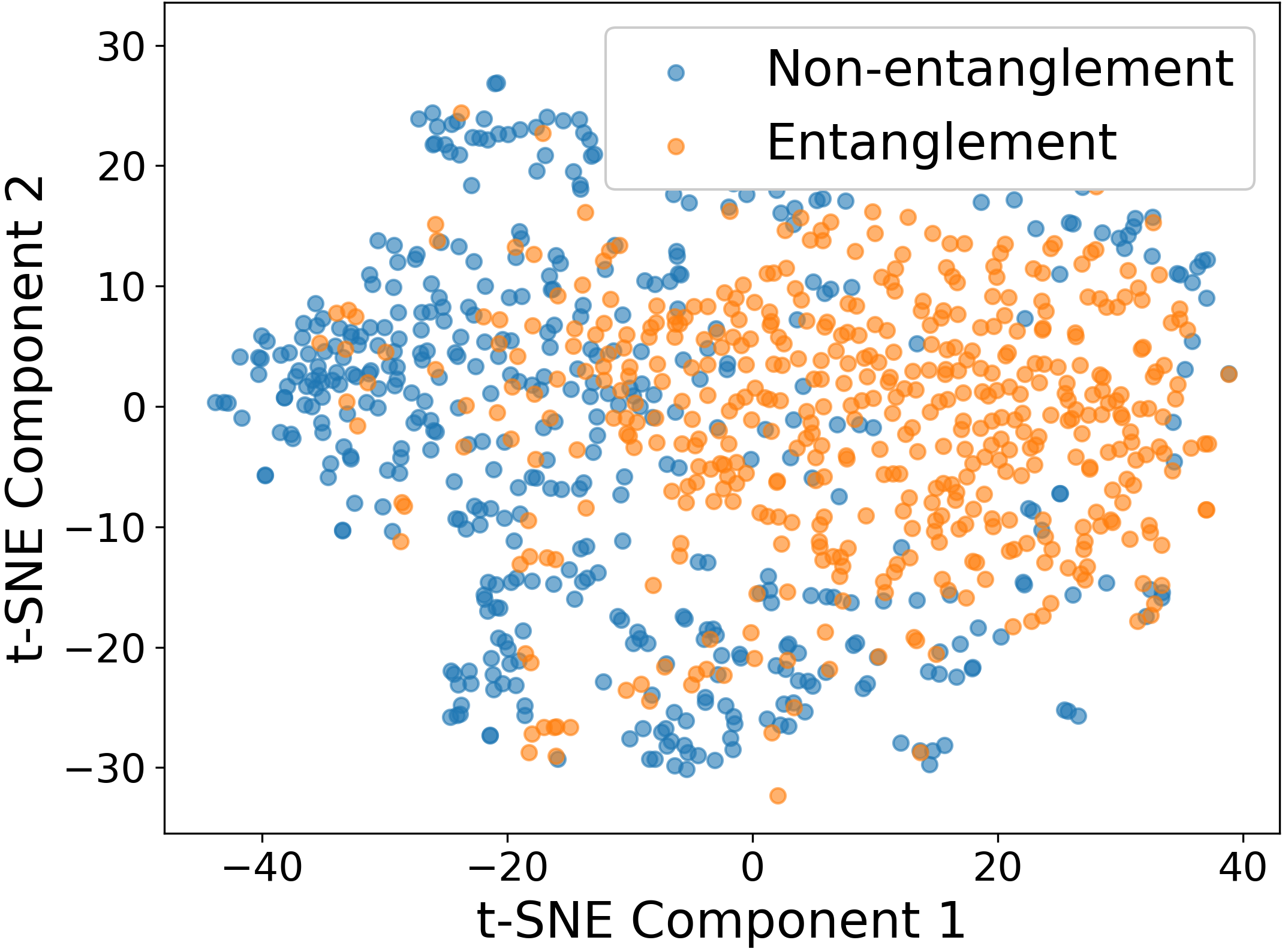}
        \caption{t-SNE - Entanglement activity}
        \label{fig:tsne_entanglement}
    \end{subfigure}
    \caption{Two-dimensional \ac{tsne} representations of the latent space for the three types of electrophysiological activity. Embeddings for rotational activity and entanglement were obtained from bipolar EGMs and for focal activity obtained from unipolar EGMs. Each point represents a sample, colored by class label: blue indicates no activity detected, and orange indicates activity detected. For each task, 1,000 randomly selected samples are shown, consisting of 500 with activity and 500 without activity.}
    \label{fig:tsne}
\end{figure}

This subsection reports the performance of the downstream classification tasks across the three driver types. We evaluated the classifiers using standard metrics and assessed latent space separability through \ac{tsne} projections. Quantitative results are presented in Table~\ref{tab:metrics}, and the corresponding \ac{tsne} visualizations are shown in Figure~\ref{fig:tsne}.

For the rotational activity classification task, CatBoost achieved the highest \ac{auc} (0.734), indicating the best overall discriminative ability. LightGBM followed closely (AUC = 0.732) and obtained the highest specificity (0.679) and accuracy (0.679), suggesting balanced performance. In contrast, KNN achieved the highest sensitivity (0.754) but suffered from low specificity (0.503) and accuracy (0.506), reflecting a tendency to overpredict the positive class. Overall, the classification models exhibited moderate performance, consistent with the substantial class overlap observed in the \ac{tsne} projections.

For the focal activity classification task, the model performance for this task was similar to the rotational case, underscoring the difficulty of discriminating focal activity. CatBoost again yielded the highest \ac{auc} (0.761), along with moderate specificity (0.734) and accuracy (0.721). KNN achieved the highest accuracy (0.751) but with low sensitivity (0.572), suggesting poor detection of positive cases. The \ac{tsne} visualization revealed a high degree of class intermixing, consistent with the moderate classifier performance.

The entanglement activity classification task showed substantially better performance across all models. CatBoost reached the highest \ac{auc} (0.928) and demonstrated strong, balanced performance with sensitivity (0.840), specificity (0.852), and accuracy (0.843). Other models also performed well, with several achieving \ac{auc}s above 0.92. These results indicate a more clearly defined latent structure for entanglement detection, as further evidenced by the \ac{tsne} visualization, which showed a marked separation between classes.

To assess real-time applicability, we evaluated the end-to-end classification time using the best-performing model (CatBoost) for the pipeline presented in Figure~\ref{fig:workflow}D. The average processing time to classify a 30-second signal was measured over 2,731 signals from 100 patients. Results showed a mean inference time of 0.0064 seconds per signal, confirming that the pipeline supports real-time classification.

\begin{table}[ht]
\centering
\begin{adjustbox}{width=\textwidth}
\begin{tabular}{lcccc cccc cccc}
\toprule
\multirow{2}{*}{\textbf{Model}} 
& \multicolumn{4}{c}{\textbf{Rotors}} 
& \multicolumn{4}{c}{\textbf{Focal}} 
& \multicolumn{4}{c}{\textbf{Entanglement}} \\
\cmidrule(lr){2-5} \cmidrule(lr){6-9} \cmidrule(lr){10-13}
& \ac{auc} & Sens. & Spec. & Acc. 
& \ac{auc} & Sens. & Spec. & Acc. 
& \ac{auc} & Sens. & Spec. & Acc. \\
\midrule
CatBoost   & \textbf{0.734} & 0.678 & 0.660 & 0.660 & \textbf{0.761} & \textbf{0.665} & 0.734 & 0.721 & \textbf{0.928} & \textbf{0.840} & 0.852 & \textbf{0.843} \\
\ac{knn}        & 0.701 & \textbf{0.754} & 0.503 & 0.506 & 0.741 & 0.572 & \textbf{0.791} & \textbf{0.751} & 0.917 & 0.824 & 0.853 & 0.830 \\
\ac{lr}         & 0.716 & 0.696 & 0.568 & 0.570 & 0.735 & 0.630 & 0.714 & 0.699 & 0.868 & 0.774 & 0.785 & 0.777 \\
LightGBM   & 0.732 & 0.678 & \textbf{0.679} & \textbf{0.679} & 0.759 & 0.659 & 0.726 & 0.714 & 0.927 & 0.828 & \textbf{0.860} & 0.835 \\
XGBoost    & 0.720 & 0.681 & 0.654 & 0.654 & 0.757 & 0.654 & 0.734 & 0.720 & 0.926 & 0.838 & 0.848 & 0.840 \\
\bottomrule
\end{tabular}
\end{adjustbox}
\caption{Performance metrics of five \ac{ml} models on the test set across three binary classification tasks. Metrics reported are \ac{auc}, Sensitivity (Sens.), Specificity (Spec.), and Accuracy (Acc.).}
\label{tab:metrics}
\end{table}

\section{Discussion}

This study evaluated the use of convolutional autoencoders to extract low-dimensional representations of intracardiac \ac{egms} recorded during atrial fibrillation. The results demonstrate that autoencoders can accurately reconstruct both unipolar and bipolar signals while enabling downstream classification of arrhythmic driver activity. In this section, we discuss the implications of these findings, address the main limitations of the approach, and outline directions for future research.

\subsection{Pipeline performance}

The results show that autoencoders trained separately on unipolar and bipolar \ac{egms} can effectively compress and reconstruct the signals, achieving low reconstruction errors and preserving morphological features. Both models achieved stable training and validation losses during the training process and demonstrated strong generalization performance in previously unseen patients, obtaining \ac{mse}s of 0.0062 in bipolar \ac{egms} and 0.0118 in unipolar \ac{egms}. The latent representations obtained are compact and informative, compressing high-dimensional inputs to 16-dimensional embeddings while retaining meaningful signal characteristics. By operating in an unsupervised manner, the autoencoder scales effectively to large datasets without requiring manual labeling. Such low-dimensional representations can serve as a basis for more efficient signal analysis, helping downstream tasks like classification, clustering, and visualization, and potentially supporting current and future ablation strategies in AF as understanding of the condition continues to evolve \cite{van20242024}.

Despite the success of dimensionality reduction and reconstruction, the classification results show variability in the discriminative capacity of embeddings in different arrhythmogenic mechanisms. CatBoost emerged as the top-performing model across all three classification tasks.

For rotor and focal activity detection, performance was moderate, with the best models achieving \ac{auc}s between 0.73 and 0.76, reflecting limited class separability. This suggests that the latent features may not fully capture the subtle signal patterns characteristic of these mechanisms. A contributing factor may be the windowing strategy, which segments the signal into fixed 1-second intervals. Since the labels are based on patterns that span multiple beats (Section~\ref{Subsec:database}), this segmentation can fragment relevant information, limiting the classifier’s ability to capture the full temporal context of each mechanism. Compared to supervised methods like those of Alhusseini \cite{alhusseini2020machine} and Liao \cite{liao2021deep}, which report \ac{auc}s above 0.90 using curated inputs or annotated features, our unsupervised approach shows more modest performance. However, it avoids reliance on manual labels or predefined structures, highlighting a trade-off between annotation-dependent accuracy and the scalability of unsupervised learning.

In contrast, the classification of entanglement yielded significantly better results, with \ac{auc}s exceeding 0.92 and consistently high accuracy, sensitivity, and specificity. These findings suggest that the embeddings obtained are particularly effective in capturing the spatial and temporal characteristics relevant to the entanglement.

The autoencoder is designed to learn global morphological patterns across channels and time, which makes it particularly effective at capturing entanglement, as this mechanism involves consistent and widespread activity across multiple electrodes. In contrast, focal activity is brief and highly localized to independent unipolar activations per single channel, making it harder for the CAE multichannel input model to capture in its latent representation. Rotational activity is more spatially distributed than focal sources but depends on subtle ``head meets tail patterns" that are not always easy to capture.

\begin{figure}[t]
    \centering
    \includegraphics[width=1\textwidth]{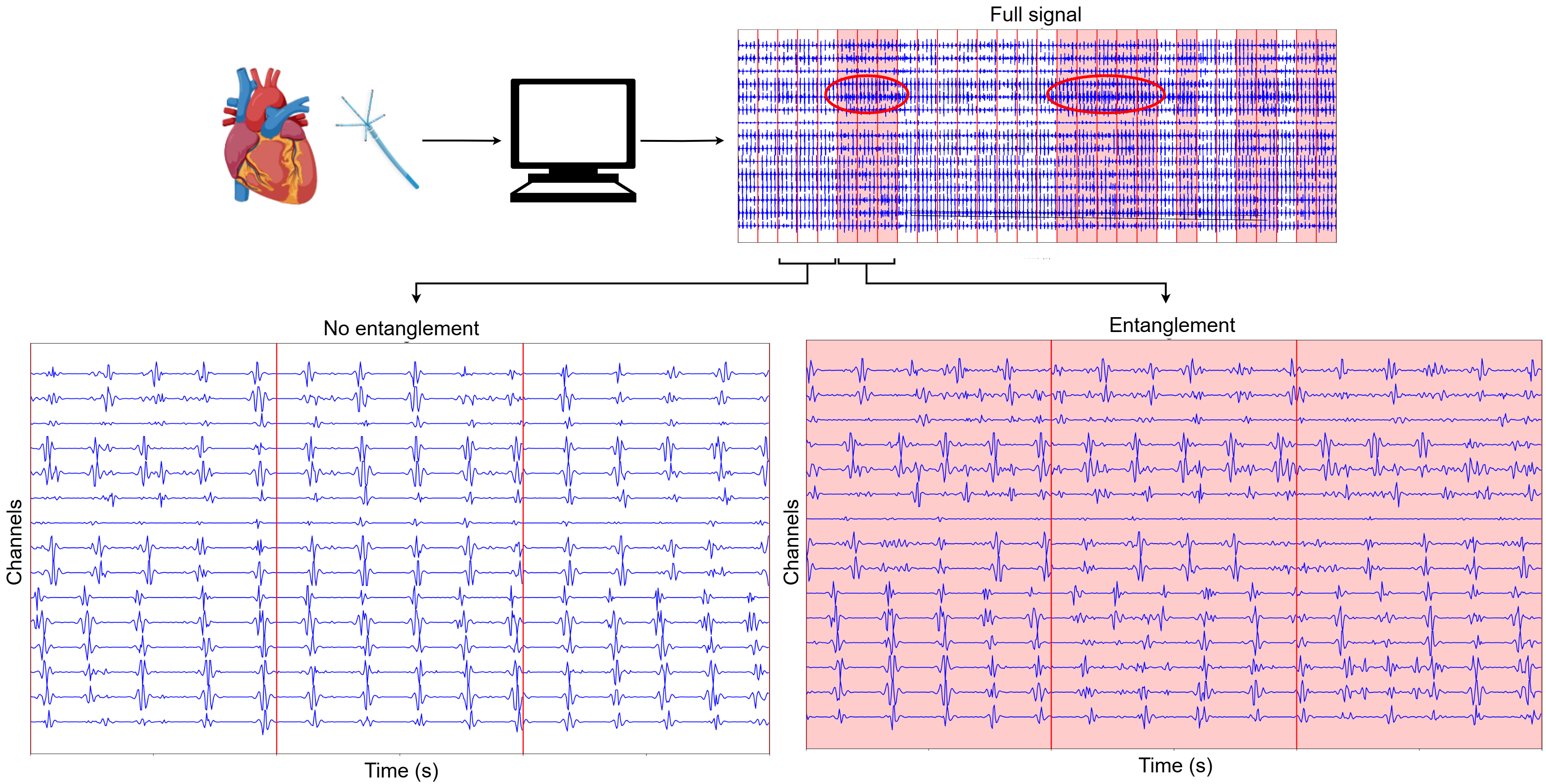}
    \caption{Application of the study's pipeline to detect entanglement in bipolar \ac{egms}. The plot shows bipolar electrograms windows, with the highlighted red regions indicating time windows labeled as containing entanglement events.}
    \label{fig:application}
\end{figure}

This entire pipeline, considering its short classification time that enables real-time performance, is well-suited for potential real-world implementation during ablation procedures. By integrating these models into platforms, such as
CARTO3 (Biosense Webster, Diamond Bar, CA, USA), EnSite Precision\textsuperscript{\textregistered} (Abbott Park, IL, USA), or Rhythmia\textsuperscript{\textregistered} (Boston Scientific Corporation, Marlborough, MA, USA), arrhythmogenic regions could be located during signal acquisition in ablation studies. Figure~\ref{fig:application} illustrates how a signal is windowed and indicates the segments where abnormal activity is detected, serving as an example of the practical application of the algorithm.  However, the final decision must always be made by the cardiologist, as this algorithm is intended solely to propose potential target regions.

\subsection{Limitations}

One of the key limitations that contribute to the differences in classification performance described above is the way labels were obtained. The annotations were generated using algorithmic methods rather than manual expert annotation. Although automated labeling helps large-scale analysis, it may miss or be inaccurate in certain events. In particular, algorithms tend to prioritize safety by favoring conservative detections, resulting in a high rate of false negatives \cite{rios2022convolutional}. This limitation could directly affect classification performance, as the model may learn from imperfect labels. Future work should consider the use of experts' manually annotated datasets to provide a more reliable ground truth that could improve the training process.

A further potential limitation, depending on the mapping catheter used in the procedures, is that the autoencoder in this study was trained with signals of 20 channels for unipolar \ac{egms} and 15 channels for bipolar \ac{egms}, based on the use of the PentaRay catheter. This architecture is not supported for catheters with a different number of channels. Training separate models with the appropriate signals for each catheter would be an appropriate solution.

Additionally, although a primary advantage of autoencoders is their ability to learn representations from unlabeled data, training the model solely to minimize reconstruction loss does not explicitly enforce class separability in the latent space. This may explain the reduced performance in tasks that involve more complex patterns. Integrating supervised or semi-supervised objectives during training could enhance the discriminative structure of the learned embeddings for downstream classification tasks.

Another important limitation relates to the interpretability of the latent space learned by the autoencoder. While the low-dimensional embeddings preserve signal morphology and enable reasonable classification, the latent space lacks structure and homogeneity. Standard autoencoders do not impose constraints on the organization of latent variables, which can lead to irregular clustering and make it difficult to associate latent dimensions with physiologically significant features. This also limits the ability to explore the space in a principled way, such as interpolating between signal types or identifying clinically relevant axes. Future work could address this by adopting variational autoencoders (VAEs), which impose a prior distribution to promote continuity and structure, potentially enhancing both interpretability and classification performance.

\section{Conclusion}

This study demonstrates the potential of \ac{cae}s for unsupervised feature extraction from unipolar and bipolar electrograms in the context of \ac{af}. The proposed pipeline effectively compresses high-dimensional \ac{egms} into compact latent embeddings that preserve key electrophysiological features, enabling accurate signal reconstruction and real-time classification of arrhythmogenic patterns, including rotational, focal, and entanglement activity.

\ac{cae}s trained separately on unipolar and bipolar \ac{egms} achieved low reconstruction errors and demonstrated strong generalization, confirming their ability to retain clinically relevant signal morphology. While classification performance for rotors and focal sources remained moderate—reflecting the intrinsic complexity and overlap of these patterns—entanglement activity was identified with high accuracy (AUC = 0.92), underscoring the discriminative power of the learned representations.

The system’s real-time inference capability further supports its integration into clinical workflows, offering a promising tool for driver localization during ablation procedures. These findings highlight the value of incorporating unsupervised deep learning into electroanatomical mapping systems to enhance data interpretation and support clinical decision-making in AF treatment.

\section*{Ethics Statement}
This study was approved by the Local Ethics Committee of the institution and conducted in accordance with applicable European and Spanish regulations on research involving human subjects. Given the retrospective nature of the study, the Ethics Committee waived the requirement for written informed consent.

\section*{Data accessibility}

All code in this study, including the \ac{cae} architecture, is openly accessible via GitHub: \href{https://github.com/PpeiroUC3M/FeatureExtractionEGMs}{https://github.com/PpeiroUC3M/FeatureExtractionEGMs}.

\section*{Declaration of AI use}

Generative artificial intelligence tools, specifically OpenAI's ChatGPT, were utilized for correcting grammar, enhancing style, rephrasing ideas, and debugging code. After using this tool, the author reviewed and edited the content as needed and assumed full responsibility for the content of the publication.

\section*{Authors' contributions}
P.P.-C.: conceptualization, data curation, formal analysis, investigation, methodology, software, writing original draft;
L.L.: investigation, methodology, writing review and editing;
P.A.: data curation, validation;
A.C.-B.: data curation, validation;
A.A.: data curation, validation;
C.S.-S.: conceptualization, resources, funding acquisition, methodology, supervision, writing review and editing;
G.R.R.-M: conceptualization, resources, funding acquisition, investigation, methodology, software, supervision, writing original draft;

\section*{Conflicts of interest}

G.R.R.-M and A.A. are co-inventors of a pending patent application related to the results presented in this article. The authors declare that the research was carried out without any commercial or financial affiliations that could be interpreted as a potential conflict of interest.

\section*{Funding}
This work was supported by the Instituto de Salud Carlos III (Madrid-Spain) (PI22-01619), and by the Madrid Government (Comunidad de Madrid-Spain) under the Multiannual Agreement with UC3M (FLAMA-CM-UC3M), and through project MAGERIT-CM (TEC-2024/COM-44).


\bibliographystyle{elsarticle-num} 
\bibliography{bibliography}

\end{document}